\pdfoutput=1
\documentclass[letterpaper, 10 pt, conference]{ieeeconf}

\IEEEoverridecommandlockouts

\usepackage{amsmath}
\usepackage{amssymb}
\usepackage{graphicx}
\usepackage{algorithm}
\usepackage{algorithmic}
\usepackage{hyperref}
\usepackage{url}
\usepackage{booktabs}
\usepackage{multirow}
\usepackage{makecell}
\usepackage{xcolor}
\usepackage{subcaption}
\usepackage{wrapfig}
\usepackage{pifont}
\newcommand{\cmark}{\ding{51}}
\usepackage[noadjust]{cite}

\title{\LARGE \bf
RoboPRO: A Perturbation-Resilient and Obstacle-Aware Benchmark for Robotic Manipulation
}
\author{Zhiyuan Li$^{1,2\dagger*}$, Charles Eret$^{1,3\dagger*}$, Dingsheng (Jack) Liu$^{1,2\dagger}$, Xuan Zhao$^{1}$, Joshua Choi$^{1,2\dagger}$,\\ Yangzheng Wu$^{1}$, Tongtong Cao$^{1}$, Yingxue Zhang$^{1}$, and Amir Rasouli$^{1,**}$%
\thanks{$*$ Equal Contribution. $^{1}$Huawei Technologies. $^{2}$University of Toronto. $^{3}$University of Waterloo}%
\thanks{$^{**}$Corresponding author: {\tt\small amir.rasouli@huawei.com}}%
\thanks{$^{\dagger}$Work done during internship at Huawei Canada.}%
}

\begin{document}

\maketitle
\thispagestyle{empty}
\pagestyle{empty}

\begin{figure*}[h!]
    \centering
    \includegraphics[width=1\textwidth]{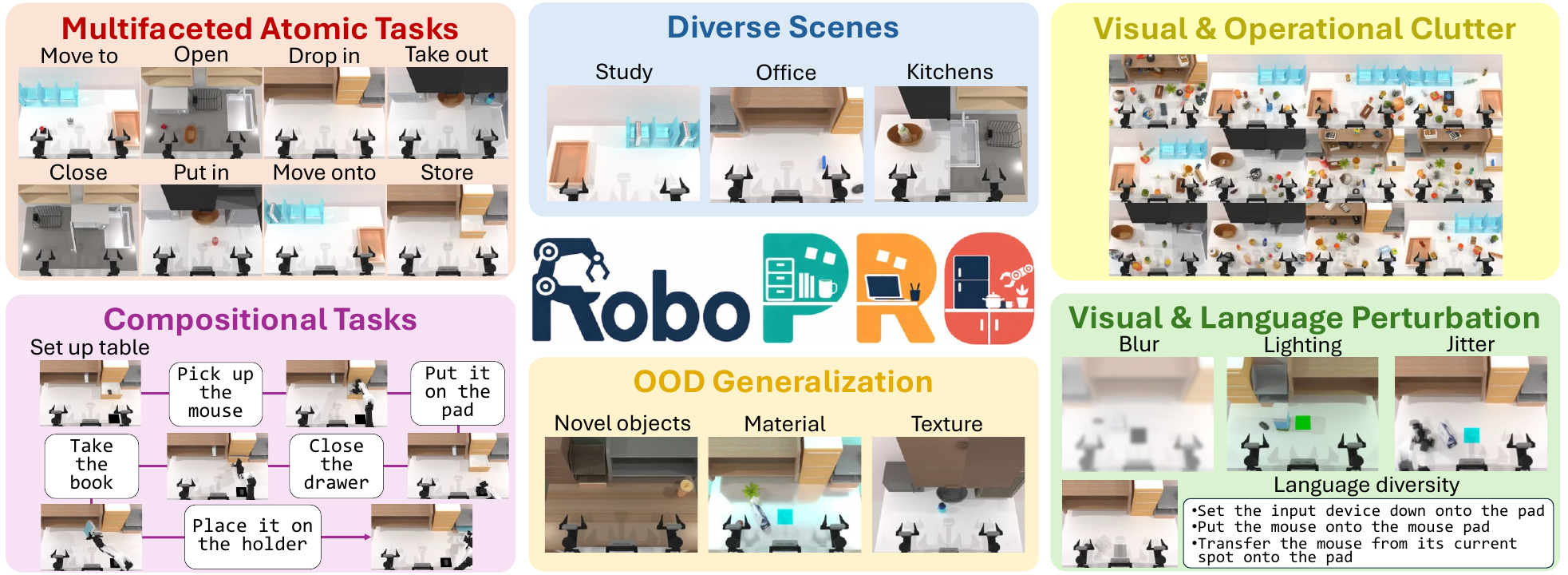}
    \caption{\textbf{RoboPRO} is a holistic safety manipulation benchmark that offers a wide range of atomic and compositional tasks in diverse scenes for evaluation of policies under visual and operational clutter, out-of-domain (OOD) distribution, and visual/language perturbations.}
    \label{fig:teaser}
    \vspace{-0.5cm}
\end{figure*}

\begin{abstract}
Object manipulation in the real world requires robotic policies to be robust to contextual perturbations and capable of safe planning in cluttered scenes. However, the majority of existing robotics benchmarks contain clean tabletop environments with success rate as the sole metric, leaving two critical deployment questions unanswered: does the policy manipulate \emph{safely} under real conditions and what does lead to its failure? Some benchmarks evaluate isolated perturbations on uncluttered surfaces or use collision measures that do not capture contact severity. Others provide realistic environments but omit perturbation evaluation and safety metrics entirely. To this end, we present \textbf{RoboPRO}, a \textbf{Robo}tics benchmark for \textbf{P}erturbation-\textbf{R}esilient and \textbf{O}bstacle-aware planning, which unifies various axes of robustness in manipulation. RoboPRO comprises 80 diverse bimanual atomic and compositional tasks across four furnished indoor environments, with a total of 16,000 demonstrations and 5.6K diverse language instructions. The scenes are populated with different numbers of distractors (non-targets), ranging from clean, involving only target objects, to densely populated to impose both visual and operational clutter. Moreover, we propose two safety-aware event metrics to determine unsafe contact between the robot and the environment or distractors. Our evaluation protocol consists of a per-factor analysis of three key perturbations, namely visual, linguistic, and object-level, with different levels of distractor densities. We conduct extensive experiments on RoboPRO using state-of-the-art manipulation policies to analyze the impact of visual and operational clutter, perturbations, and instructional diversity. Our findings indicate clutter can drastically reduce success and increase non-target collision, different perturbations have different effects on the performance, and that training data composition affects overall performance.  Our findings highlight the key challenges for manipulation in real-world conditions and motivate future research in this domain. The benchmark suite and data are available at \url{https://robopro-bench.github.io/RoboPRO/}.

\end{abstract}

\section{Introduction}
\label{sec:introduction}

Learning-based robotic manipulation policies have demonstrated impressive generalization on tabletop tasks in simulation~\cite{liu2023libero,zhang2025vlabench}, yet deploying these policies in complex environments exposes two distinct failure modes that current benchmarks largely overlook. The first is \emph{perturbation fragility}: policies that perform near-ceiling on the canonical evaluation regress sharply when camera pose, lighting, language phrasing, or object identity drift even modestly. The second is \emph{collision behaviour}: a policy that completes its task by disturbing a non-target object or forcefully pushing through a cabinet door achieves a perfect score on the standard success rate (SR) metric even though it is unsafe to deploy \cite{rasouli2026howvlas}. Despite rapid progress in policy learning and evaluation, the existing benchmarks do not jointly stress both axes within the same set of obstacle-laden scenes.

Perturbation-robustness suites~\cite{pumacay2024colosseum,fei25libero-plus,sedlacek2025realm} omit the effect of camera vibration and bundle language factors, so individual perturbation factors are understudied. In addition, they report only success rate (SR) on clean tabletops with minimal surrounding geometry, so the contact events that would expose a brittle planner never occur. Conversely, the safety-oriented benchmarks that do instrument collision~\cite{aegis2024safelibero,roboeval2025,vlaarena2025} use binary or count-based contact flags that conflate a 2\,mm cup nudge with a sweep that knocks the cup off the table. In these benchmarks, the scene setups lack realistic, complex background, have limited diversity, and clutter perturbation is not systematic.

To this end, we propose RoboPRO, a safety benchmark that evaluates perturbation-resilience and obstacle-awareness of manipulation policies under a unified suite (see Fig.~\ref{fig:teaser}). RoboPRO consists of 60 atomic tasks and 20 compositional tasks performed in four distinct,  realistically furnished simulated household environments under diverse configurations. Our benchmark extends upon existing perturbation protocols by modeling pixel shift, full OOD furniture texture and object diversification, and decomposes language perturbations along five axes, enabling more in-depth analysis of instruction diversity on performance. Moreover, we employ a clutter generation strategy to induce manipulation under different degrees of visual and operational complexity. To measure safety, we propose two novel metrics to determine hazardous contact between the robot and environment or distractors. In sum, our contributions are as follows: 1) \textbf{A holistic safety benchmark} that combines perturbation- and clutter-based challenges into a unified evaluation suite using diverse scenes, and atomic and compositional manipulation tasks; 2) \textbf{Extended perturbation strategy} for capturing real-world camera movement and per-axis analysis for measuring the impact of instruction diversity on performance; 3) \textbf{A cluttering strategy} allowing systematic scenario generation with different degrees of visual and operational complexity; 4) \textbf{Safety metrics} to quantify the hazardous actions of the robot with respect to the environment or non-target objects; 5) \textbf{Extensive evaluation}  of state-of-the-art manipulation policies on RoboPRO, with in-depth analysis of the effect of perturbations and clutter, training data composition, and task complexity on performance.

\section{Related Work}
\label{sec:related_work}
\subsection{Robotic Action Policies}
Emerging action policies fall into a few architectural categories that differ in action parameterization and model size: flow-matching, which integrates a continuous trajectory along a learned ODE ~\cite{wang2025xvla,black2024pi0,intelligence2025pi05}; autoregressive, which decodes actions as discrete tokens~\cite{kim2024openvla,brohan2022rt1,brohan2023rt2}; diffusion, which denoises action chunks ~\cite{holobrain2026,li2024cogact,octo_RSS_2024}. To probe whether perturbation and collision behavior varies with architectural choice, we evaluate top-performing open models with different sizes, namely $\pi$0~\cite{black2024pi0}, $\pi$0.5~\cite{intelligence2025pi05}, and X-VLA~\cite{wang2025xvla}.

\subsection{Perturbation, Clutter, and Policy Robustness}
Despite achieving strong performance on a variety of complex tasks, learning-based robotic policies are shown to be vulnerable under real-world conditions. A number of studies~\cite{pumacay2024colosseum,mu2025robotwin} show visual perturbations, such as lighting, surface or background texture, or camera pose, can greatly hamper performance but at different rates. Similarly, language perturbations have also been shown to reduce performance, although less so compared to visual perturbations because the policies rely more on visual input than language instructions~\cite{fei25libero-plus}. Visual and operational clutter due to the presence of distractors (non-target objects) can also significantly lower policy success~\cite{rasouli2026distracted,pakdamansavoji2026nice,dipalo2024kat,tang2023selective} by causing object confusion, grasping failure, and collisions. RoboPRO is designed to address the aforementioned  problems in a unified framework that provides a systematic strategy for generating scenarios encompassing various perturbation and clutter-related challenges.

\subsection{Robotic Manipulation Benchmarks}
\noindent\textbf{Scenario design.}
Existing manipulation benchmarks fall into two broad categories: \emph{tabletop / single-skill suites} that exercise a clean per-skill evaluation ~\cite{liu2023libero,rlbench,fei25libero-plus}. In these benchmarks, the environment is often simplified and contains mainly the target objects and a handful of distractors with limited diversity. \emph{Household-scale suites} model realistic environments populated with furniture and diverse objects with long-horizon coverage ~\cite{nasiriany2026robocasa365,li2024behavior1k}. In these suites the focus is on skill learning, hence they contain no systematic perturbations or clutter beyond task-relevant objects. RoboPRO sits between these two categories. It inherits the furnished, multi-skill, bimanual character of household suites while applying perturbations at evaluation time \emph{within} cluttered scenes.

\noindent\textbf{Robustness evaluation.}
For robustness evaluation, some benchmarks apply visual perturbations, such as lighting, surface texture \cite{mu2025robotwin}, camera pose~\cite{pumacay2024colosseum,roboviewbias2025}, and sensor noise via blurring and contrast shift~\cite{sedlacek2025realm}. LIBERO-Plus~\cite{fei25libero-plus} also adds language perturbation via adding distracting words, switching objects with common sense descriptions, and changing instruction complexity. RoboPRO combines these factors and extends the perturbation by altering furniture textures and material which may affect manipulation in confined spaces, adding pixel shift to capture real-world platform vibration, and expanding language perturbations into five axes, and performs per-factor analysis as opposed to holistic evaluation in~\cite{fei25libero-plus}. Scene clutter by distractors has also been included in many benchmarks. However, the clutter either merely serves as visual distraction outside of the operational space of the robot~\cite{mu2025robotwin,pumacay2024colosseum,sedlacek2025realm} or is formed sparsely with a small number of objects limiting its impact on the generated path~\cite{aegis2024safelibero,vlaarena2025}. RoboPRO studies the combined effect of perturbation and clutter by offering a systematic strategy for clutter generation with diverse in-distribution and OOD objects within the canonical trajectory's swept volume, emphasizing both visual and operational challenges. Our benchmark contains both large-scale  clean and cluttered environments accompanied by an evaluation protocol for examining different degrees of clutter. Table~\ref{tab:benchmark_summary} provides a comparison against the closest manipulation benchmarks.

\begin{table}[t]
\centering
\caption{\textbf{Manipulation benchmarks.} \cmark: supported; blank: not explicitly reported under the column definition. Under \emph{Perturbations},  systematic \textbf{language}, \textbf{visual}, and \textbf{object} perturbations applied at evaluation time. Under \emph{Clutter}, \textbf{Visual}: distractor objects on the workspace surface (visual clutter only); \textbf{Op.\ clut.}: obstacles in the manipulation path.}
\label{tab:benchmark_summary}
\small
\setlength{\tabcolsep}{4pt}
\resizebox{1\columnwidth}{!}{
\begin{tabular}{@{}lcccccc@{}}
\toprule
 & & \multicolumn{3}{c}{Perturbations} & \multicolumn{2}{c}{Clutter} \\
\cmidrule(lr){3-5}\cmidrule(lr){6-7}
Benchmark & Tasks & Language & Visual & Object & Visual clut.& Operational clut. \\ \midrule
LIBERO~\cite{liu2023libero}                       & 130   &        &        &        &        &  \\
LIBERO-Plus~\cite{fei25libero-plus}               & 40    & \cmark & \cmark &        & \cmark & \\
LIBERO-Safety~\cite{cui2026liberosafety}          & 75 & \cmark & \cmark & \cmark &        & \cmark \\
Colosseum~\cite{pumacay2024colosseum}             & 20    &        & \cmark & \cmark & \cmark &  \\
RoboTwin 2.0~\cite{chen2025robotwin2}             & 50    & \cmark & \cmark &        & \cmark &        \\
VLA-Arena~\cite{vlaarena2025}                     & 170   & \cmark & \cmark & \cmark & \cmark &        \\
RoboCasa365~\cite{nasiriany2026robocasa365}       & 365   &        &        &        &        &  \\
BEHAVIOR-1K~\cite{li2024behavior1k}               & 1000  &        &        &        &        &  \\
SIMPLER~\cite{li2024simpler}                      & 8     &        & \cmark & \cmark & \cmark &        \\
GenieSim 3.0~\cite{yin2026geniesim3}              & 200+  & \cmark & \cmark & \cmark &        &        \\ \midrule
\textbf{RoboPRO (Ours)}                            & 80    & \cmark & \cmark & \cmark & \cmark & \cmark \\
\bottomrule
\end{tabular}}\vspace{-0.5cm}
\end{table}

\noindent\textbf{Metrics.}
Robotic policies are mainly evaluated using success rate, i.e. the ratio of the tasks the robot successfully completed. However, as a recent study~\cite{rasouli2026howvlas} suggests, in realistic household environments, when safety constraints are considered, the true success can be substantially lower. Works on manipulation safety~\cite{roboeval2025,safevla2025,ye2025ra,yang2026cape,aegis2024safelibero} additionally report on collision rate. However, they treat any form of contact as collision, which can greatly exaggerate actual hazardous behavior since a certain degree of contact is acceptable, or even necessary in the course of manipulation. In RoboPRO, we offer two new metrics which measure collision based on significance of change in the status of non-target objects and force between the robot's arm and the environment.

\section{Benchmark Setup}
\label{sec:benchmark}

\subsection{Problem Formulation}
\label{sec:problem_formulation}

We consider language-conditioned manipulation in partially observable environments. A task is specified by
\[
\mathcal{T}
=
(\mathcal{S},\mathcal{A},\mathcal{O},P,\rho_0,\ell,g,T),
\]
where $\mathcal{S}$, $\mathcal{A}$, and $\mathcal{O}$ denote the state, action, and observation spaces; $P(s_{t+1}\mid s_t,a_t)$ defines the transition dynamics; and $\rho_0$ specifies the initial-state distribution, including target placement and obstacle layout. The instruction $\ell$ specifies the desired manipulation, $g:\mathcal{S}\rightarrow\{0,1\}$ defines task success, and $T$ is the episode horizon. A policy $\pi_\theta(a_t\mid o_{\leq t},\ell)$ selects actions from the observation history and instruction, with observations comprising RGB images and robot proprioception.

At each clutter level, we evaluate the policy under visual, object, and language perturbations, denoted by $\phi=(\phi_v,\phi_o,\phi_\ell)$ (Section~\ref{sec:perturbations}), with $\phi=\mathbf{0}$ denoting the unperturbed condition. For each episode, $G=g(s_T)\in\{0,1\}$ indicates task success at the terminal state, and $C=\mathcal{E}_{\mathrm{O}}\lor\mathcal{E}_{\mathrm{F}}\in\{0,1\}$ indicates whether an object-displacement or furniture-contact event occurs. The event criteria are specified in Section~\ref{sec:experimental_setup}.

We evaluate collision-free task completion through
\begin{equation}
    J(\pi_\theta;\phi)
    = \mathbb{E}_{\pi_\theta,\phi}\!\left[G(1-C)\right],
    \label{eq:objective}
\end{equation}
where the expectation is over sampled tasks, initial states, and policy rollouts at the specified clutter level. This quantity is the population hard success rate (HSR). We also report success rate and collision rate, corresponding to $\mathbb{E}[G]$ and $\mathbb{E}[C]$, respectively. Their empirical estimates use the episode-level observations $G^{(i)}$ and $C^{(i)}$.

\subsection{Environments and Tasks}
\label{sec:env_tasks}

\begin{figure*}[t]
    \centering
    \includegraphics[width=0.8\textwidth]{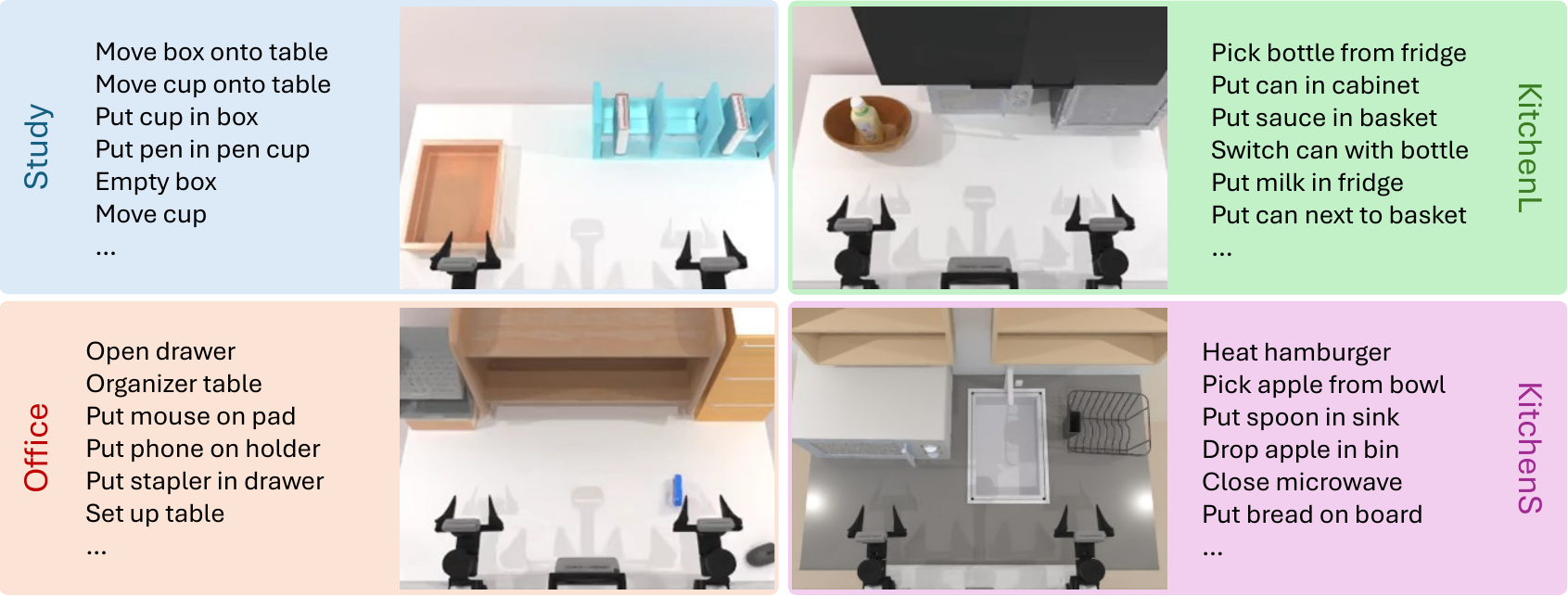}
    \caption{\textbf{Task examples across environments.} Representative tasks from each of the four environments, illustrating the diversity of manipulation skills exercised by the benchmark.}
    \label{fig:tasks}   \vspace{-0.5cm}
\end{figure*}

RoboPRO contains four environments spanning functionally distinct workspaces---\textbf{Office} (desk, drawers, shelf), \textbf{Study} (table, bookcase, storage box), \textbf{Kitchen-S} (countertop, sink, microwave, overhead shelves), and \textbf{Kitchen-L} (refrigerator, deep cabinets, storage baskets) (see Fig.~\ref{fig:tasks}). Each environment configuration is randomly selected from three predefined layouts.
 All scenes are built in the SAPIEN simulator~\cite{Xiang_2020_SAPIEN} with a dual-arm Aloha-Agilex robot. In aggregate, the four scenes contribute 29 manipulable object types and 16 fixed furniture pieces with one articulated part each (drawers, cabinet, microwave, fridge), and tasks split evenly with 15 atomic and 5 compositional per scene (60+20 total).

RoboPRO comprises 80 tasks across the four indoor environments.
Within each environment, the tasks are organized into two complexity tiers: \emph{atomic} tasks contain a single primitive skill (e.g., \texttt{pick\_apple\_from\_sink}, \texttt{put\_mouse\_on\_pad}), while \emph{compositional} tasks chain 2--4 skills in a multi-step sequence (e.g., open a cabinet, retrieve an item, close the cabinet, place it on a shelf). Representative tasks are shown in Fig.~\ref{fig:tasks}.

\subsection{Clutter Generation}
\label{sec:obstacles}
In RoboPRO we generate both clean environments containing only the furniture and task-relevant objects and cluttered environments populated with distracting obstacles. A distractor is a non-target object placed deliberately to constrain (or obscure) the feasible motion. Distracting obstacles are never the goal of an instruction, never grasped, and contribute only to the geometric and collision-related component of the task. The separation of distractors from target objects matters for evaluation: a successful policy reaches and manipulates the \emph{target object} while avoiding contact with all \emph{distractors} (and with fixed furniture, which forms a third category).

\begin{figure*}[t]
    \centering
    \includegraphics[width=\textwidth]{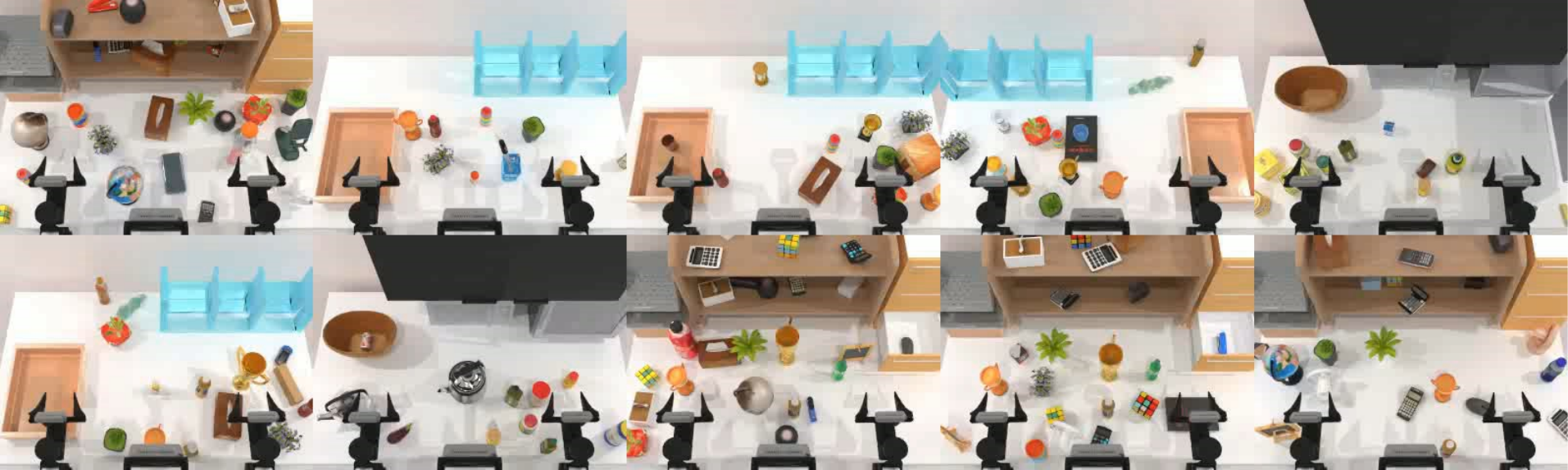}
    \caption{Examples of \textbf{cluttered scenes} with different placements and obstacle density.}
    \label{fig:obstacle_density}
\end{figure*}

We populate the workspace with obstacles drawn from a pool of items unique to each scene in two height categories: \emph{short} obstacles, e.g. calculators, tissue boxes, staplers and \emph{tall} obstacles, e.g. bottles, plant pots, trophies. In each episode, $k$ obstacles are sampled uniformly from the full pool with a ratio of 30/70 (short/tall) to impose more challenge.  Obstacles are positioned by a procedural placement routine that samples poses within the workspace volume swept by the canonical demonstration trajectory, with a minimum inter-object spacing of 5\,cm to prevent overlap. This sampling rule guarantees that obstacles intersect the operational path of a successful trajectory, requiring active avoidance rather than mere visual filtering (see Fig.~\ref{fig:obstacle_density}).

\subsection{Dataset}
\label{sec:dataset}

RoboPRO provides 16{,}000 demonstration trajectories, consisting of 100 \emph{clean} trajectories (no distractors) and 100 \emph{cluttered} trajectories (with obstacle density $k$ sampled from $\{6,\ldots,15\}$, yielding 10 episodes per density level) for each of the 80 tasks. All demonstrations are collected via scripted motion-planning policies using cuRobo \cite{sundaralingam2023curobo} for collision-aware trajectory optimization. To stress instruction-following without conflating it with the harder evaluation-time language axes, we additionally pair each canonical instruction with a \emph{training-time} pool of 20 synonymous instructions per task that preserves action / object / target / constraint semantics; GPT5.4 generated variants, which were reviewed by humans, yielding $20 \times 80 = 1{,}600$ training paraphrases. The held-out evaluation pool that targets harder semantic axes is described in Section~\ref{sec:perturbations}.

\subsection{Inference-time Perturbations}
\label{sec:perturbations}

\begin{figure*}[t]
    \centering
    \includegraphics[width=\textwidth]{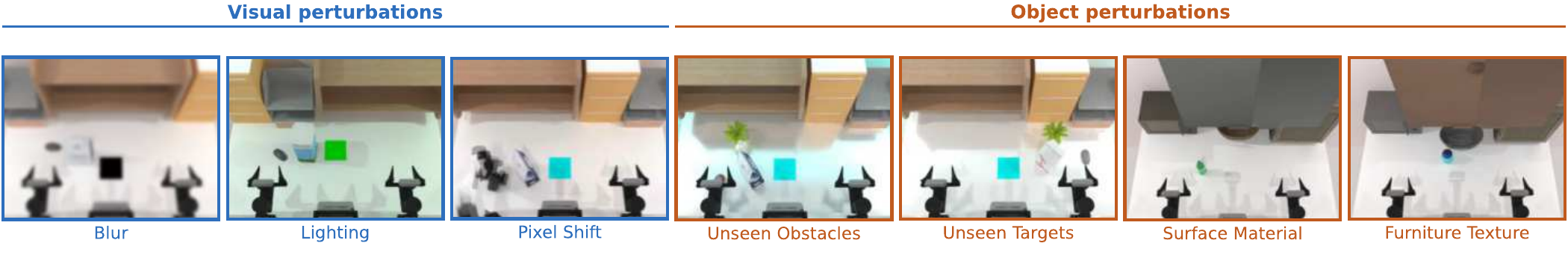}\vspace{-0.3cm}
    \caption{Examples of the \textbf{visual perturbation }axes (blue) and \textbf{object-level perturbation} axes (orange), annotated per panel.}
    \label{fig:perturbations}    \vspace{-0.5cm}
\end{figure*}

\paragraph{Visual perturbations.} We use the following three vision axes (see Fig.~\ref{fig:perturbations}): \emph{Lighting} samples a per-episode color multiplier (default range $[0.4, 1.8]$) and toggles directional lighting and shadow casting; \emph{Blur} applies per-frame Gaussian blur with a fixed kernel $\sigma=0.5$; \emph{Pixel Shift} (jitter) applies a per-frame random integer translation of up to $5$ pixels. Lighting is fixed once at scene reset; blur and Pixel Shift are resampled every frame, so all three stack independently with the OOD axes below.

\paragraph{Language perturbations.} For each task we generate a 50-variant evaluation pool, disjoint from the training-time synonym pool (see Section~\ref{sec:dataset}), that stresses surface complexity along five more challenging axes inspired by~\cite{fei25libero-plus}: (1)~\emph{distractor context}---irrelevant background, such as time of day, mood, or narrative; (2)~\emph{multi-object distraction}---references to other plausible objects in the scene while still uniquely identifying the target; (3)~\emph{commonsense indirect}---referring to the object or target by a functional or categorical property rather than its direct name (``the baked item'' for bread); (4)~\emph{reasoning chain}---a preamble or goal explanation that requires inference to recover the action; and (5)~\emph{colloquial / polite}---fillers, hedges, and request-form constructions. GPT5.4 generated the variants, which were  validated through human review. Ten validated variants per axis yield 50 evaluation perturbations per task, or $50 \times 80 = 4{,}000$ held-out instructions across the benchmark. Because the five evaluation axes are \emph{disjoint} from the four light-weight axes used at training time, robust performance requires generalization beyond the surface forms seen during fine-tuning rather than memorization.

\paragraph{Object-level perturbations.} To examine whether  policies rely on appearance memorization or identity rather than on spatial structure, we consider the following shifts (see Fig.~\ref{fig:perturbations})--- \emph{appearance}: \emph{surface material} resamples table physically-based rendering (PBR) parameters per episode (metallic $\in [0, 0.8]$, roughness $\in [0.05, 0.95]$), and \emph{furniture texture}  (e.g., shelves, cabinets, fridge) by drawing from a held-out pool of $1{,}000$ images per object class. We also consider shifts in \emph{scene composition} by including \emph{unseen obstacles} and \emph{unseen targets} drawn  from a held-out object pool. Spatial layout, robot pose, and task goal remain fixed across all four axes; only geometry and appearance change. The axes share no state and can be enabled in any combination.

\section{Experiments}
\label{sec:experiments}
In this section, our goal is to evaluate state-of-the-art robotic manipulation policies using the proposed benchmark, RoboPRO. Specifically, we aim to answer the following research questions: \textbf{Q1--How does scene clutter affect task success?} \textbf{Q2--How does training data composition affect clean and cluttered performance?} \textbf{Q3--How robust are the existing policies to various contextual perturbations?}  \textbf{Q4--How do visual, linguistic, and object-level perturbations affect performance?}  \textbf{Q5--How does task compositionality affect task success and safety?} \textbf{Q6--How does performance vary with clutter density?}

\subsection{Experimental Setup}
\label{sec:experimental_setup}

\paragraph{Training data.} Unless otherwise specified, all models are trained on the full RoboPRO training set of $200 \times 80 = 16{,}000$ demonstrations: 200 trajectories per task, consisting of 100 clean and 100 cluttered scenes. For cluttered scenes we consider density $k$ sampled uniformly from $\{6, \ldots, 15\}$ for each of the 80 tasks. To encourage generalization and prevent overfitting, each task instance is paired with one of 20 paraphrased instructions (drawn from the simple-difficulty pool described in Section~\ref{sec:dataset}) and randomly initialized object positions and instances.%

\textbf{Baselines.} We evaluate three vision-language-action (VLA) policies, namely \textbf{$\pi$0}, \textbf{$\pi$0.5}, and \textbf{X-VLA}. All models are jointly fine-tuned on all tasks on the same per-task training split using their respective default hyperparameters.

\textbf{Evaluation Protocol.} Each model is evaluated for 20 episodes per task in the clean setting. In the cluttered setting, we evaluate two episodes at each distractor density, yielding 20 cluttered episodes per task. Clean and cluttered aggregates therefore contain equal numbers of evaluation episodes.

\textbf{Evaluation Metrics.}
\label{sec:metrics}
We report three Bernoulli-mean metrics over $N$ episodes---success rate (SR), collision rate (CR), and hard success rate (HSR), i.e. SR with no collision:
\begin{equation}
\begin{split}
    \text{SR} = \tfrac{1}{N}\!\sum_{i} G^{(i)},\quad
\text{CR} = \tfrac{1}{N}\!\sum_{i} C^{(i)},\quad \\
\text{HSR} = \tfrac{1}{N}\!\sum_{i} G^{(i)}\bigl(1 - C^{(i)}\bigr),
\end{split}
\label{eq:metrics_main}
\end{equation}
where $G^{(i)}$ and $C^{(i)}$ are the binary success and collision indicators for episode $i$, as defined in Section~\ref{sec:problem_formulation}. For collision events, we propose \textbf{two new metrics} as $C^{(i)}= \mathcal{E}_{\text{O}}^{(i)} \lor \mathcal{E}_{\text{F}}^{(i)}$. $\mathcal{E}_{\text{O}}^{(i)}$ is the object collision event measured by changes in the state of any non-target object ($>$1\,cm in position or $>$0.1\,rad in orientation set empirically). $\mathcal{E}_{\text{F}}^{(i)}$ captures the furniture event, i.e. hazardous contact with the environment triggered by $>$10\,N sustained contact force on fixed furniture. We refer to the CR of objects and furniture as OCR and FCR, respectively.

\subsection{Experimental Results}
\begin{table*}[t]
    \centering
    \caption{
        \textbf{Success rate in clean vs.\ cluttered scenes.}
        Per-environment performance with no perturbations applied.
        Each cell reports \textbf{SR\,/\,HSR}.
        Results are averaged within each environment:
        \textbf{Off} (Office), \textbf{Stu} (Study),
        \textbf{K-S} (Kitchen-S), and \textbf{K-L} (Kitchen-L).
        Higher is better for all metrics.
    }
    \label{tab:main_results_clutter}
    \footnotesize
    \setlength{\tabcolsep}{2pt}

    \resizebox{\textwidth}{!}{%
        \begin{tabular}{@{}l ccccc | ccccc | c@{}}
            \toprule
            & \multicolumn{5}{c|}{\textbf{Clean (SR/HSR)}}
            & \multicolumn{5}{c|}{\textbf{Cluttered (SR/HSR)}}
            & \multirow{2}{*}{\textbf{Overall SR/HSR}} \\
            \cmidrule(lr){2-6}
            \cmidrule(lr){7-11}

            Model
            & Off & Stu & K-S & K-L & \emph{Avg.}
            & Off & Stu & K-S & K-L & \emph{Avg.}
            & \\

            \midrule

            X-VLA
            & 0.33/0.32
            & 0.69/0.67
            & 0.22/0.20
            & \textbf{0.75}/\textbf{0.68}
            & 0.50/0.47
            & 0.27/0.24
            & 0.57/0.44
            & 0.11/0.09
            & \textbf{0.65}/0.33
            & 0.40/0.28
            & 0.45/0.37 \\

            $\pi_0$
            & 0.39/0.38
            & 0.74/0.71
            & 0.61/0.56
            & 0.66/0.59
            & 0.60/0.56
            & 0.26/0.21
            & 0.58/0.43
            & 0.45/0.37
            & 0.61/0.32
            & 0.47/0.33
            & 0.54/0.44 \\

            $\pi_{0.5}$
            & \textbf{0.56}/\textbf{0.54}
            & \textbf{0.85}/\textbf{0.83}
            & \textbf{0.68}/\textbf{0.62}
            & 0.73/0.64
            & \textbf{0.70}/\textbf{0.66}
            & \textbf{0.48}/\textbf{0.40}
            & \textbf{0.68}/\textbf{0.57}
            & \textbf{0.63}/\textbf{0.56}
            & \textbf{0.65}/\textbf{0.39}
            & \textbf{0.61}/\textbf{0.48}
            & \textbf{0.66}/\textbf{0.57} \\

            \bottomrule
        \end{tabular}%
    }
    \vspace{-0.5cm}
\end{table*}

\paragraph{R1--Performance is decreased in cluttered scenes leading to safety hazards.} Table~\ref{tab:main_results_clutter} summarizes SR and HSR across all four environments for clean and cluttered conditions. $\pi$0.5 achieves the highest SR at 0.70 and 0.61 in clean and cluttered scenes, respectively.
All three policies exhibit lower success rates under clutter, with absolute drops of 0.13, 0.10, and 0.09 for $\pi_0$, X-VLA, and $\pi_{0.5}$, respectively. The degradation likely reflects multiple effects of clutter, including collisions, grasp failures, and visual confusion, rather than collisions alone. This motivates examining safety-aware metrics separately, as SR alone does not reveal whether failures arise from unsafe contact or other errors.

SR alone does not capture the safety cost of clutter. Across all three models, HSR is consistently lower than SR, indicating that a non-negligible fraction of otherwise successful trajectories involve unsafe contact. Under cluttered conditions, $\pi$0, X-VLA, and $\pi$0.5 achieve SRs of 0.47, 0.40, and 0.61, respectively, while their HSRs decrease to 0.33, 0.28, and 0.48. This gap highlights that evaluating task completion alone can overestimate policy performance in cluttered environments. Notably, $\pi$0.5 retains most of its relative advantage over X-VLA under the safety-aware HSR metric, suggesting that its higher task success does not come at the expense of increased unsafe contact.

\begin{figure}
    \centering
\includegraphics[width=0.9\linewidth]{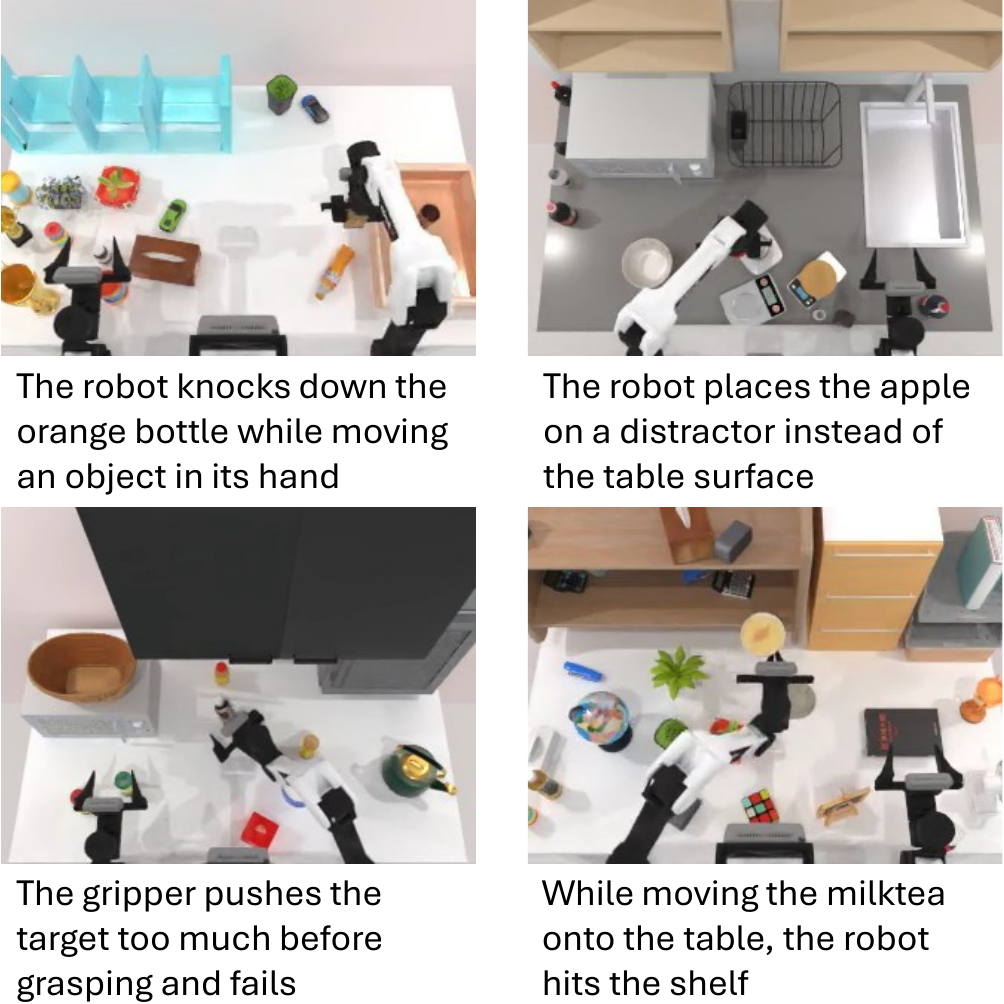}
    \caption{Qualitative examples of policies' failures in scenes with different degrees of clutter.}\vspace{-0.5cm}
    \label{fig:qual_examples}
\end{figure}

Table~\ref{tab:main_collision_results} further shows that collisions are not exclusive to cluttered scenes. Even in clean environments, policies can contact furniture and secondary task-related objects, resulting in CRs between 0.09 and 0.15. Introducing distractors substantially increases collision rates for all models, with CR rising to 0.45 for $\pi_0$, 0.39 for X-VLA, and 0.32 for $\pi_{0.5}$. This increase is driven primarily by object collisions: OCR rises from 0.03 in clean scenes to 0.37, 0.35, and 0.27, respectively, whereas FCR remains approximately unchanged or slightly decreases. These results suggest that the dominant safety challenge introduced by clutter is interaction with non-target objects along the manipulation path, rather than increased contact with fixed furniture. Among the evaluated policies, $\pi$0.5 achieves both the highest cluttered-scene SR and the lowest overall CR, while X-VLA exhibits the lowest furniture collision rate. Qualitative examples of these failure modes are shown in Fig.~\ref{fig:qual_examples}.

 \begin{table}[t]
\centering
\caption{\textbf{Collision in clean vs.\ cluttered scenes.}  Results are averaged over all scenes with no perturbations. For all results lower is better.}
\label{tab:main_collision_results}
\footnotesize
\setlength{\tabcolsep}{4pt}
\resizebox{0.8\columnwidth}{!}{%
\begin{tabular}{@{}l ccc | ccc }
\toprule
& \multicolumn{3}{c|}{\textbf{Clean }} & \multicolumn{3}{c}{\textbf{Cluttered}} \\
\cmidrule(lr){2-4}\cmidrule(lr){5-7}
Model & FCR&OCR& CR & FCR&OCR& CR  \\
\midrule
X-VLA    & 0.07 & 0.03 & 0.09 & 0.06 & 0.35 & 0.39 \\
$\pi$0   & 0.12 & 0.03 & 0.15 & 0.12 & 0.37 & 0.45 \\
$\pi$0.5 & 0.09 & 0.03 & 0.12 & 0.08 & 0.27 & 0.32 \\
\bottomrule
\end{tabular}%
 }
 \vspace{-0.2cm}
 \end{table}

\begin{table}[t]
\centering
\caption{\textbf{Effect of training data composition} on $\pi$0.5, evaluated on the median-12 tasks using combinations of clean and cluttered data. For all results higher is better.}
\label{tab:ablation_data}
\footnotesize
\setlength{\tabcolsep}{4pt}
\resizebox{0.8\columnwidth}{!}{
\begin{tabular}{@{}lcccc@{}}
\toprule
Training split & Cl.SR & Clt.SR & Clt.HSR & Ovr.SR \\
\midrule
Clean only  & 0.712 & 0.238 & 0.158 & 0.475 \\
Clutter only & 0.667 & \textbf{0.554} & \textbf{0.383} & 0.611 \\
Half-split             & 0.779 & 0.502 & 0.360 & 0.641 \\
Default   & \textbf{0.787} & 0.513 & 0.337 & \textbf{0.650} \\
\bottomrule
\end{tabular}}
\vspace{-0.5cm}
 \end{table}

\paragraph{R2--Training on a mix of clean and cluttered data yields the best overall performance.} We select $\pi$0.5, the model with the highest overall SR, and train it on four subsets of the data: \emph{Default} containing the entire 16K training set, \emph{half-split} containing half of the clean and half of the cluttered scenes, and \emph{clean-only} and \emph{clutter-only}. The latter three subsets contain 8K samples each. All four models are evaluated on 12 tasks selected near the median baseline SR (three per environment), with results reported in Table~\ref{tab:ablation_data}. The best overall performance is achieved with the default set as expected. However, half-split trails Default by only 0.009 overall SR, within the uncertainty of the evaluation. This indicates that the models already converge well at 8K samples, and it grounds the performance differences among the three subsets in training data composition rather than incomplete training. Among the 8K subsets, half-split achieves the highest overall SR, supporting that mixed training data yields the best performance.

The two single-regime models show distinct behaviors. Clutter-only leads all models, including Default, in cluttered SR and HSR. Its training data consist entirely of collision-aware demonstrations, so the model never has to reconcile avoidance behavior with the unconstrained motions of clean demonstrations. It also generalizes to clean scenes, which it never saw during training, reaching an SR of 0.667. By contrast, clean-only degrades the most under clutter, reaching less than half the cluttered SR and HSR of the other models.
The mixed training sets achieve the strongest overall SR, whereas clutter-only training produces the highest cluttered SR and HSR. This indicates a trade-off between broad performance and specialization for cluttered conditions.

\begin{table}[t]
    \centering
    \caption{
        \textbf{Perturbations impact.}
        Results are averaged over median-12 tasks (three from each environment)
        and over both clean and cluttered scenes, for 13 perturbation axes grouped as
        \emph{Object} (OOD appearance, unseen obstacles, unseen targets, and their combination),
        \emph{Visual} (blur, lighting, pixel shift, and their combination), and
        \emph{Language} (distractor context, multi-object distraction, commonsense indirect,
        reasoning chain, and colloquial/polite instructions).
        $\uparrow$ and $\downarrow$ indicate higher or lower is better, respectively.
        Subscripts show the delta relative to \emph{Original} (unperturbed baseline). 
    }
\label{tab:main_results_perturbation}
    \small

    \resizebox{\columnwidth}{!}{%
        \begin{tabular}{@{}ll c|cccc@{}}
            \toprule
            Model & Metric & Original & Object & Visual & Language & \textbf{Pert.\ Avg.} \\
            \midrule

            \multirow{3}{*}{X-VLA}
            & SR$\uparrow$
            & 0.41
            & $0.41_{\textcolor{gray}{\text{+0.00}}}$
            & $0.44_{\textcolor{green}{\text{+0.03}}}$
            & $0.33_{\textcolor{red}{\text{-0.08}}}$
            & $0.39_{\textcolor{red}{\text{-0.02}}}$ \\

            & HSR$\uparrow$
            & 0.35
            & $0.35_{\textcolor{gray}{\text{+0.00}}}$
            & $0.36_{\textcolor{green}{\text{+0.01}}}$
            & $0.29_{\textcolor{red}{\text{-0.06}}}$
            & $0.33_{\textcolor{red}{\text{-0.02}}}$ \\

            & CR$\downarrow$
            & 0.26
            & $0.25_{\textcolor{green}{\text{-0.01}}}$
            & $0.30_{\textcolor{red}{\text{+0.04}}}$
            & $0.27_{\textcolor{red}{\text{+0.01}}}$
            & $0.28_{\textcolor{red}{\text{+0.02}}}$ \\

            \midrule

            \multirow{3}{*}{$\pi_0$}
            & SR$\uparrow$
            & 0.50
            & $0.49_{\textcolor{red}{\text{-0.01}}}$
            & $0.53_{\textcolor{green}{\text{+0.03}}}$
            & $0.44_{\textcolor{red}{\text{-0.06}}}$
            & $0.48_{\textcolor{red}{\text{-0.02}}}$ \\

            & HSR$\uparrow$
            & 0.43
            & $0.43_{\textcolor{gray}{\text{+0.00}}}$
            & $0.44_{\textcolor{green}{\text{+0.01}}}$
            & $0.38_{\textcolor{red}{\text{-0.05}}}$
            & $0.42_{\textcolor{red}{\text{-0.01}}}$ \\

            & CR$\downarrow$
            & 0.27
            & $0.32_{\textcolor{red}{\text{+0.05}}}$
            & $0.28_{\textcolor{red}{\text{+0.01}}}$
            & $0.32_{\textcolor{red}{\text{+0.05}}}$
            & $0.31_{\textcolor{red}{\text{+0.04}}}$ \\

            \midrule

            \multirow{3}{*}{$\pi_{0.5}$}
            & SR$\uparrow$
            & 0.65
            & $0.65_{\textcolor{gray}{\text{+0.00}}}$
            & $0.69_{\textcolor{green}{\text{+0.04}}}$
            & $0.58_{\textcolor{red}{\text{-0.07}}}$
            & $\mathbf{0.63}_{\textcolor{red}{\text{-0.02}}}$ \\

            & HSR$\uparrow$
            & 0.55
            & $0.58_{\textcolor{green}{\text{+0.03}}}$
            & $0.61_{\textcolor{green}{\text{+0.06}}}$
            & $0.53_{\textcolor{red}{\text{-0.02}}}$
            & $\mathbf{0.57}_{\textcolor{green}{\text{+0.02}}}$ \\

            & CR$\downarrow$
            & 0.24
            & $0.27_{\textcolor{red}{\text{+0.03}}}$
            & $0.24_{\textcolor{gray}{\text{+0.00}}}$
            & $0.29_{\textcolor{red}{\text{+0.05}}}$
            & $\mathbf{0.27}_{\textcolor{red}{\text{+0.03}}}$ \\
            \bottomrule
        \end{tabular}%
    }        \vspace{-0.5cm}

\end{table}

\paragraph{R3--Performance degradation under perturbations is far less compared to clutter.} Table~\ref{tab:main_results_perturbation} reports the impact of the three perturbation categories, averaged over clean and cluttered scenes. Interestingly, averaged over all perturbations, SR drops by only 0.02 for every model, which is substantially smaller than the clutter-related drop (R1). On average, the ranking $\pi$0.5 $>$ $\pi$0 $>$ X-VLA is also preserved under every perturbation category and metric although X-VLA has the lowest CR under language and object-level perturbations overall. $\pi$0.5 achieves the strongest absolute perturbed performance, but the relative drops do not consistently show that stronger models are more perturbation-resilient. Moreover, not all perturbations degrade the models. Language perturbations reduce SR by 0.06--0.08 and raise CR by up to 0.05 across models, whereas object-level perturbations barely shift the three metrics, and visual perturbations even increase SR and HSR.

\begin{figure}[t]
    \centering
    \begin{minipage}[c]{0.48\columnwidth}
        \centering
        \includegraphics[width=\linewidth]{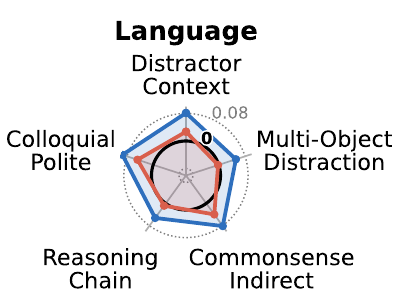}
    \end{minipage}\hfill
    \begin{minipage}[c]{0.48\columnwidth}
        \centering
        \includegraphics[width=\linewidth]{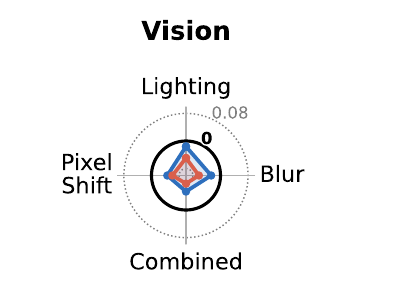}
    \end{minipage}

    \vspace{2pt}
    \begin{minipage}[c]{0.48\columnwidth}
        \centering
        \includegraphics[width=\linewidth]{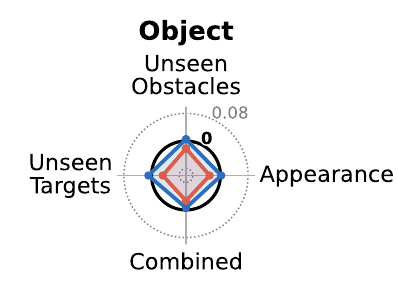}
    \end{minipage}\hfill
    \begin{minipage}[c]{0.48\columnwidth}
        \centering
        \includegraphics[width=0.95\linewidth]{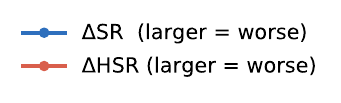}
    \end{minipage}\vspace{-0.4cm}
    \caption{\textbf{Per-axis perturbation sensitivity of $\pi$0.5.}  Polygons show the \emph{drop} from the unperturbed baseline, averaged over clean and cluttered scenes. The bold black circle marks zero and points indicate improvement over the baseline.}
\label{fig:perturbation_radars}\vspace{-0.3cm}
\end{figure}

\paragraph{R4--Perturbation impacts vary across individual axes.}
To understand why the three perturbation categories affect the models differently, Fig.~\ref{fig:perturbation_radars} plots the per-axis drop in SR and HSR for $\pi$0.5 relative to the unperturbed baseline, averaged over clean and cluttered scenes; HSR follows the same trend as SR throughout.

All five language axes cause considerable drops. Each axis replaces or removes the action and object words, or injects non-target distractors into the instruction. For example, ``put the apple on the plate'' may become ``look at both fruits on the table and place the red one on the plate while the banana stays''. Neither ``put'' nor ``apple'' appears explicitly, and in a cluttered scene full of distractors such phrasing causes misgrounding and failure. The same mechanism explains why SR barely shifts under object-level perturbations. Given a correct task description, the policy attends to the text-specified target and is minimally distracted by unseen objects. Among the object axes, unseen targets shift SR the most. The instruction still describes the original target while an unseen instance replaces it in the scene, creating the largest text--vision mismatch. Yet even this drop stays below 0.01, further suggesting that grounding is driven by the text rather than by visual appearance. The rise of CR under language perturbations further indicates that a misgrounded instruction can also lead to unsafe contact.

Visual perturbations leave SR at or slightly above the baseline. Among the single axes, pixel shift helps the most, raising SR by 0.05. Such an effect can partially be due to the evaluation variance and partially due to the fact that pixel shift may force the model to re-analyze the image and re-align the robot relative to the target throughout the episode. The remaining axes improve as well, indicating that current policies generalize to unseen visual perturbations. Overall, current policies are markedly more sensitive to how an instruction is phrased than to what the camera sees or which object instances appear.

\begin{figure}
    \vspace{-0.3cm}
    \centering
    \includegraphics[width=0.6\columnwidth]{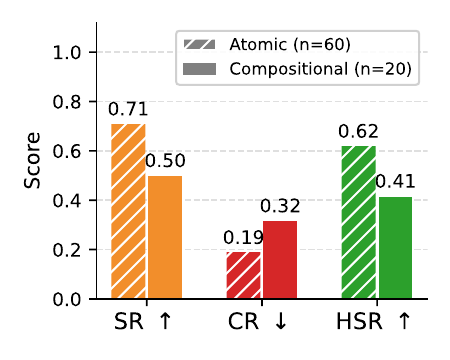}\vspace{-0.3cm}
    \caption{\textbf{Atomic vs.\ compositional task performance ($\pi$0.5),} averaged over all 80 tasks under unperturbed conditions.}
    \label{fig:atomic_vs_comp}
    \vspace{-0.5cm}
\end{figure}
\paragraph{R5--Compositional tasks have lower success and higher collision rates, with the extra collisions concentrated in failed episodes.}
Compositional tasks chain 2--4 primitive skills, exposing more sub-goals to potential contact and extending trajectory length. Fig.~\ref{fig:atomic_vs_comp} reports SR, CR, and HSR for $\pi$0.5 split by complexity tier. From atomic to compositional tasks, SR and HSR drop by 0.21. Each additional skill introduces another opportunity for failure. CR in turn rises from 0.19 to 0.32. Composing more sub-skills requires the policy to travel longer distances through densely cluttered space, increasing its exposure to unsafe contact. Comparing the three metrics further reveals where these collisions occur. SR$-$HSR measures collisions within successful episodes, and CR$-$(SR$-$HSR) measures collisions within failures. In atomic tasks, collisions split almost evenly between successes and failures (0.09 vs.\ 0.10). In compositional tasks, the success share stays at 0.09 while the failure share grows to 0.23. The additional collisions of longer tasks therefore do not come from chaining itself, but from failing episodes. As errors accumulate along the chain and the rollout drifts away from the training distribution, the policy fails the task and simultaneously fails to avoid obstacles.

\paragraph{R6--Clutter affects policies similarly, plateauing at lower densities.}
\label{app:density_trends}

\begin{figure}[h]
    \centering \vspace{-0.4cm}
    \includegraphics[width=0.9\linewidth]{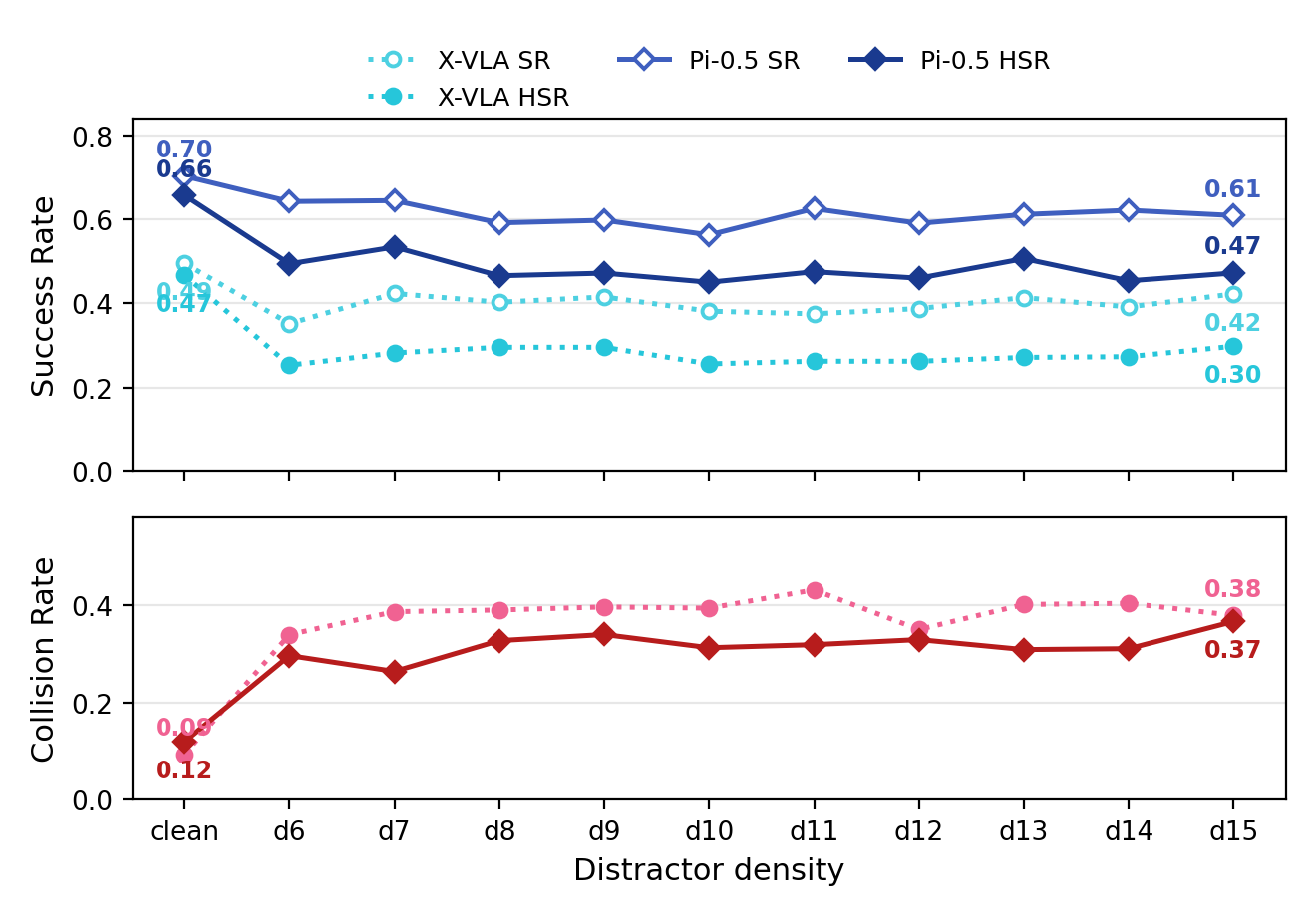}
    \caption{\textbf{Performance across distractor densities.} Per-model SR, HSR, and CR averaged over the four scenes, from clean to cluttered conditions ($k\in\{6,\ldots,15\}$).}
    \label{fig:density_trend}
    \vspace{-0.5cm}
\end{figure}

Fig.~\ref{fig:density_trend} compares $\pi_{0.5}$ and X-VLA across success rate (SR), hard success rate (HSR), and collision rate (CR) as distractor density grows from clean to $k{=}15$. The bulk of the degradation occurs  at the lowest density. From clean to $k{=}6$, HSR falls by 0.16 for $\pi_{0.5}$ and 0.21 for X-VLA, and CR roughly triples. Beyond $k{=}6$ the curves largely plateau: SR and HSR fluctuate within a narrow band while CR drifts up mildly, reaching 0.37 and 0.38 at $k{=}15$. Since two independently trained models follow the same pattern, the degradation is attributable to the benchmark's clutter design rather than any single model's weakness. This shows that a handful of distractors is sufficient to trigger most of the unsafe behavior.

Despite the shared trend, $\pi_{0.5}$ dominates X-VLA at every density and on all three metrics across $k\in\{6,\ldots,15\}$. The ranking and the margins established in the clean condition persist as density grows, indicating that clutter rescales performance without reordering the models. Notably, the stronger model is also the safer one under clutter, so task competence and collision avoidance are not in tension.

\section{Conclusion}
\label{sec:conclusion}
We proposed RoboPRO, a novel benchmark for evaluating robotic manipulation under operational clutter and contextual perturbations. RoboPRO comprises 80 atomic and compositional tasks, swept-volume obstacle placement, and contact-aware safety evaluation metrics. Our experiments indicate that clutter causes the largest performance degradation and introduces additional safety risks. Under perturbations, language variations consistently cause larger performance drops than visual and object-level perturbations in both clean and cluttered scenes. We also observe that our data diversification strategy can negate some effects of clutter.

\bibliographystyle{IEEEtran}
\bibliography{references}

\begin{thebibliography}{10}
\providecommand{\url}[1]{#1}
\csname url@rmstyle\endcsname
\providecommand{\newblock}{\relax}
\providecommand{\bibinfo}[2]{#2}
\providecommand\BIBentrySTDinterwordspacing{\spaceskip=0pt\relax}
\providecommand\BIBentryALTinterwordstretchfactor{4}
\providecommand\BIBentryALTinterwordspacing{\spaceskip=\fontdimen2\font plus
\BIBentryALTinterwordstretchfactor\fontdimen3\font minus
  \fontdimen4\font\relax}
\providecommand\BIBforeignlanguage[2]{{%
\expandafter\ifx\csname l@#1\endcsname\relax
\typeout{** WARNING: IEEEtran.bst: No hyphenation pattern has been}%
\typeout{** loaded for the language `#1'. Using the pattern for}%
\typeout{** the default language instead.}%
\else
\language=\csname l@#1\endcsname
\fi
#2}}

\bibitem{liu2023libero}
B.~Liu, Y.~Zhu, \emph{et~al.}, ``Libero: Benchmarking knowledge transfer for
  lifelong robot learning,'' in \emph{NeurIPS}, 2023.

\bibitem{zhang2025vlabench}
S.~Zhang, Z.~Xu, \emph{et~al.}, ``Vlabench: A large-scale benchmark for
  language-conditioned robotics manipulation with long-horizon reasoning
  tasks,'' in \emph{ICCV}, 2025.

\bibitem{rasouli2026howvlas}
A.~Rasouli, Y.~Wu, \emph{et~al.}, ``How {VLA}s (really) work in open-world
  environments,'' \emph{arXiv preprint arXiv:2604.21192}, 2026.

\bibitem{pumacay2024colosseum}
W.~Pumacay, I.~Singh, \emph{et~al.}, ``The colosseum: A benchmark for
  evaluating generalization for robotic manipulation,'' in \emph{RSS}, 2024.

\bibitem{fei25libero-plus}
S.~Fei, S.~Wang, J.~Shi, \emph{et~al.}, ``Libero-plus: In-depth robustness
  analysis of vision-language-action models,'' \emph{arXiv preprint
  arXiv:2510.13626}, 2025.

\bibitem{sedlacek2025realm}
M.~Sedlacek, P.~Yefanov, \emph{et~al.}, ``Realm: A real-to-sim validated
  benchmark for generalization in robotic manipulation,'' \emph{arXiv preprint
  arXiv:2512.19562}, 2025.

\bibitem{aegis2024safelibero}
S.~Hu \emph{et~al.}, ``Vlsa: Vision-language-action models with plug-and-play
  safety constraint layer,'' \emph{arXiv preprint arXiv:2512.11891}, 2025.

\bibitem{roboeval2025}
Y.~R. Wang, C.~Ung, \emph{et~al.}, ``Roboeval: Where robotic manipulation meets
  structured and scalable evaluation,'' \emph{arXiv preprint arXiv:2507.00435},
  2025.

\bibitem{vlaarena2025}
B.~Zhang, J.~Li, \emph{et~al.}, ``Vla-arena: An open-source framework for
  benchmarking vision-language-action models,'' \emph{arXiv preprint
  arXiv:2512.22539}, 2025.

\bibitem{wang2025xvla}
J.~Zheng, J.~Li, \emph{et~al.}, ``{X-VLA}: Soft-prompted transformer as
  scalable cross-embodiment vision-language-action model,'' in \emph{ICLR},
  2026.

\bibitem{black2024pi0}
K.~Black, N.~Brown, \emph{et~al.}, ``$\pi_0$: A vision-language-action flow
  model for general robot control,'' in \emph{RSS}, 2025.

\bibitem{intelligence2025pi05}
{Physical Intelligence} \emph{et~al.}, ``{$\pi_{0.5}$}: a
  vision-language-action model with open-world generalization,'' in
  \emph{CoRL}, 2025.

\bibitem{kim2024openvla}
M.~J. Kim, K.~Pertsch, \emph{et~al.}, ``{OpenVLA}: An open-source
  vision-language-action model,'' in \emph{CoRL}, 2024.

\bibitem{brohan2022rt1}
A.~Brohan \emph{et~al.}, ``{RT-1}: Robotics transformer for real-world control
  at scale,'' in \emph{RSS}, 2023.

\bibitem{brohan2023rt2}
A.~Brohan, N.~Brown, \emph{et~al.}, ``{RT-2}: Vision-language-action models
  transfer web knowledge to robotic control,'' in \emph{CoRL}, 2023.

\bibitem{holobrain2026}
X.~Lin, T.~Lin, \emph{et~al.}, ``{HoloBrain-0} technical report,'' \emph{arXiv
  preprint arXiv:2602.12062}, 2026.

\bibitem{li2024cogact}
Q.~Li, Y.~Liang, \emph{et~al.}, ``Cogact: A foundational vision-language-action
  model for synergizing cognition and action in robotic manipulation,''
  \emph{arXiv preprint arXiv:2411.19650}, 2024.

\bibitem{octo_RSS_2024}
D.~Ghosh, H.~R. Walke, \emph{et~al.}, ``Octo: An open-source generalist robot
  policy,'' in \emph{RSS}, 2024.

\bibitem{mu2025robotwin}
Y.~Mu, T.~Chen, \emph{et~al.}, ``Robotwin: Dual-arm robot benchmark with
  generative digital twins,'' in \emph{CVPR}, 2025.

\bibitem{rasouli2026distracted}
A.~Rasouli \emph{et~al.}, ``Distracted robot: How visual clutter undermine
  robotic manipulation,'' in \emph{ICRA}, 2026.

\bibitem{pakdamansavoji2026nice}
S.~Pakdamansavoji, M.~Pourkeshavarz, \emph{et~al.}, ``Improving robotic
  manipulation robustness via nice scene surgery,'' in \emph{ICRA}, 2026.

\bibitem{dipalo2024kat}
N.~Di~Palo and E.~Johns, ``Keypoint action tokens enable in-context imitation
  learning in robotics,'' in \emph{RSS}, 2024.

\bibitem{tang2023selective}
B.~Tang and G.~S. Sukhatme, ``Selective object rearrangement in clutter,'' in
  \emph{CoRL}, 2023.

\bibitem{rlbench}
S.~James, Z.~Ma, \emph{et~al.}, ``Rlbench: The robot learning benchmark \&
  learning environment,'' \emph{RAL}, vol.~5, no.~2, pp. 3019--3026, 2020.

\bibitem{nasiriany2026robocasa365}
S.~Nasiriany, S.~Nasiriany, A.~Maddukuri, and Y.~Zhu, ``Robocasa365: A
  large-scale simulation framework for training and benchmarking generalist
  robots,'' in \emph{ICLR}, 2026.

\bibitem{li2024behavior1k}
C.~Li, R.~Zhang, J.~Wong, \emph{et~al.}, ``Behavior-1k: A benchmark for
  embodied ai with 1,000 everyday activities and realistic simulation,'' in
  \emph{CoRL}, 2022.

\bibitem{roboviewbias2025}
E.~Liu, S.~Liang, \emph{et~al.}, ``Roboview-bias: How camera viewpoint
  amplifies visual bias in robotic manipulation,'' \emph{arXiv preprint
  arXiv:2509.22356}, 2025.

\bibitem{cui2026liberosafety}
R.~Cui, Z.~Zhang, \emph{et~al.}, ``{LIBERO-Safety}: A comprehensive benchmark
  for physical and semantic safety in vision-language-action models,'' in
  \emph{ECCV}, 2026.

\bibitem{chen2025robotwin2}
T.~Chen, Z.~Chen, \emph{et~al.}, ``Robotwin 2.0: A scalable data generator and
  benchmark with strong domain randomization for robust bimanual robotic
  manipulation,'' \emph{arXiv preprint arXiv:2506.18088}, 2025.

\bibitem{li2024simpler}
X.~Li, K.~Hsu, J.~Gu, \emph{et~al.}, ``Evaluating real-world robot manipulation
  policies in simulation,'' in \emph{CoRL}, 2024.

\bibitem{yin2026geniesim3}
C.~Yin, D.~Huang, \emph{et~al.}, ``Genie sim 3.0: A high-fidelity comprehensive
  simulation platform for humanoid robot,'' \emph{arXiv preprint
  arXiv:2601.02078}, 2026.

\bibitem{safevla2025}
B.~Zhang \emph{et~al.}, ``Safevla: Towards safety alignment of
  vision-language-action model via constrained learning,'' in \emph{NeurIPS},
  2025.

\bibitem{ye2025ra}
X.~Ye, R.~H. Yang, \emph{et~al.}, ``Ra-dp: Rapid adaptive diffusion policy for
  training-free high-frequency robotics replanning,'' in \emph{IROS}, 2025.

\bibitem{yang2026cape}
R.~H. Yang, X.~Zhao, \emph{et~al.}, ``Cape: Context-aware diffusion policy via
  proximal mode expansion for collision avoidance,'' \emph{arXiv preprint
  arXiv:2511.22773}, 2025.

\bibitem{Xiang_2020_SAPIEN}
F.~Xiang, Y.~Qin, \emph{et~al.}, ``{SAPIEN}: A simulated part-based interactive
  environment,'' in \emph{CVPR}, 2020.

\bibitem{sundaralingam2023curobo}
B.~Sundaralingam \emph{et~al.}, ``Curobo: Parallelized collision-free robot
  motion generation,'' in \emph{ICRA}, 2023.

\end{thebibliography}

\end{document}